\documentclass{article} 
\usepackage{iclr2027_conference,times}

\usepackage{amsmath,amssymb}
\usepackage{microtype}
\usepackage{graphicx}
\usepackage{booktabs}
\usepackage{array}
\usepackage[table]{xcolor}
\usepackage{float}
\usepackage{needspace}
\usepackage{placeins}
\usepackage{etoc}
\usepackage[hidelinks,linktoc=all]{hyperref}
\usepackage{url}
\usepackage{cleveref}

\definecolor{tablegroup}{RGB}{236,238,240}
\definecolor{tablerule}{RGB}{160,166,173}
\definecolor{cigray}{RGB}{95,100,108}
\definecolor{tablegray}{RGB}{239,242,245}
\definecolor{rulegray}{RGB}{165,172,180}
\definecolor{TableHeader}{RGB}{242,244,247}
\definecolor{TablePanel}{RGB}{247,248,250}
\definecolor{TableBase}{RGB}{250,250,250}
\definecolor{TablePooled}{RGB}{246,249,253}
\definecolor{TableHighlight}{RGB}{232,241,252}
\definecolor{TableHighlightStrong}{RGB}{219,233,250}
\definecolor{tableaccent}{HTML}{EDF2F6}

\newcommand{\estci}[3]{%
  $#1$\,{\scriptsize\color{cigray}$[#2,\,#3]$}%
}

\newcommand{\SplitSizes}{The split holds 305 sequences on GSM8K, 313 on MATH and 317 on commonsense; the Mistral--MATH NLL scoring audit contains 312 sequences.}
\newcommand{\Ntraj}{75}
\newcommand{\Nconfig}{24}
\newcommand{\Nckpt}{1038}

\newcommand{\LastRegret}{0.97}

\newcommand{\SpreadMed}{3.5}

\newcommand{\NovelGenNll}{+0.74}

\newcommand{\NovelAgrNll}{+0.90}

\newcommand{\NovelOracleIn}{+1.33}
\newcommand{\NovelOracleOut}{+0.05}

\newcommand{\NovelOracleGap}{+1.29}

\newcommand{\NovelAgrGap}{+0.31}

\title{From Checkpoint Variation to Selection Gains\\in Supervised Fine-Tuning}

\author{%
   Yupeng Chang\textsuperscript{1}, Wenxuan Zhang\textsuperscript{3}, Yuan Wu\textsuperscript{1,2}\thanks{Corresponding author} \\
   \textsuperscript{1}School of Artificial Intelligence, Jilin University\\
   \textsuperscript{2}Key Laboratory of Symbolic Computation and Knowledge Engineering, Jilin University\\
   \textsuperscript{3}Singapore University of Technology and Design\\
   {\small\texttt{changyp23@mails.jlu.edu.cn, wxzhang@sutd.edu.sg, yuanwu@jlu.edu.cn}} \\
}

\iclrfinalcopy

\begin{document}

\maketitle
\lhead{}
\etocdepthtag.toc{main}

\begin{abstract}
Checkpoint selection is a routine decision in supervised fine-tuning (SFT):
training produces multiple checkpoints, but only one is retained. Yet
fixed-budget comparisons do not by themselves distinguish three empirical claims: whether
more validation data improve checkpoint selection, whether a selection rule
outperforms validation-loss selection, and whether it improves over simply
retaining the final checkpoint. We therefore treat checkpoint selection as a
finite-information decision problem: the decision must be made using limited
validation data. Holding completed training trajectories, candidate
checkpoints, and independent test items fixed, we vary the validation budget
and separately measure improvement from additional validation data, gain over
negative log-likelihood (NLL) selection, and gain over the final checkpoint.
Across 60 mathematical SFT trajectories and 19 configurations, increasing the
validation budget from 32 to 305--313 examples raises independent-test accuracy
by $+0.32$ percentage points (pp) for generated-accuracy selection and
$+0.29$ pp for checkpoint agreement, with 95\% configuration-bootstrap CIs
of $[+0.10,+0.56]$ and $[+0.11,+0.50]$, respectively. At the full validation
budget, the two generation-based rules outperform matched NLL selection by
$+0.71$ and $+0.85$ pp, respectively, yet their gains over the final
checkpoint remain unresolved. A cross-domain replication on 12 newly trained
Commonsense trajectories, with its protocol frozen before training and checkpoint
choices frozen before independent-test evaluation, shows the same qualitative
separation: increasing the validation budget from 32 to 1,024 questions
improves generated-accuracy and checkpoint-agreement selection by $+0.87$
and $+0.27$ pp, while their gains over the final checkpoint again remain
unresolved. Together, these results show that benefiting from more validation
data, outperforming NLL selection, and outperforming the final checkpoint are
distinct empirical claims that require separate evidence.
\par\smallskip
{\centering
\textbf{Code and artifacts:} \href{https://github.com/llm172/checkpoint-selection-reproduction}{\textcolor{blue}{GitHub repository}}
\par}
\end{abstract}

\section{Introduction}
\label{sec:intro}

Supervised fine-tuning (SFT) produces a sequence of model checkpoints
during training, but only one is typically retained for evaluation or
deployment. \emph{Checkpoint selection} is the decision of which checkpoint
to keep from a completed training trajectory. In practice, selection commonly
relies on validation loss or held-out generation quality; a simple alternative
is to retain the final checkpoint. The choice can matter substantially: in the
original 75-trajectory grid, the median observed best-to-worst accuracy spread
within a trajectory is $\SpreadMed$ percentage points (pp).

Yet observable checkpoint variation does not imply that a better checkpoint
can be identified reliably. With finite validation data, the checkpoint that
looks best on the selection set may partly reflect sampling variation.
Moreover, outperforming validation-loss selection does not establish that the
selected checkpoint improves over the simpler decision of retaining the final
checkpoint. This creates an evidence gap: variation visible along a training
trajectory is not the same as variation that finite validation data can
reliably convert into better performance on new examples. This distinction
also matters in practice because validation is itself a resource: larger
validation sets require additional scoring and, for label-based selection,
additional labels or reference answers.

We therefore treat checkpoint selection as a \emph{finite-information decision
problem}: the decision must be made using limited validation data. Rather than
asking only which rule performs best at a single validation size, we
distinguish three questions: whether more validation data improve selection,
whether a rule outperforms validation-loss selection, and whether it improves
over simply retaining the final checkpoint. We vary the validation budget and
measure how the selected checkpoint performs on independent test items as more
validation data become available. We call this relationship the
\emph{validation-budget response}. Rather than proposing a new
checkpoint-selection algorithm, we measure the value of additional validation
data while holding the underlying training trajectory fixed.

\begin{figure}[t]
    \centering
    \includegraphics[width=.98\linewidth]{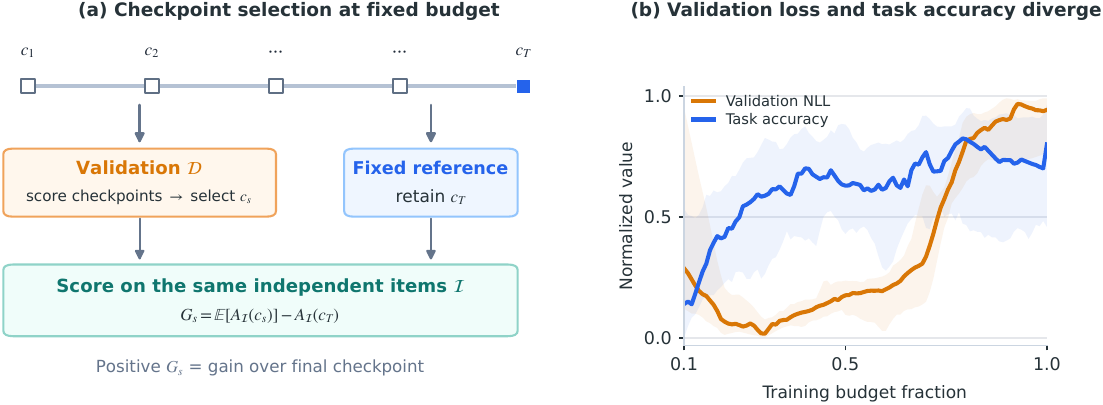}
\caption{
\textbf{Checkpoint variation and attainable selection gain are distinct.}
\textbf{(a)} For a completed SFT trajectory, a validation-based rule
selects checkpoint $c_s$, while the fixed reference retains the final
checkpoint $c_T$. Both decisions are evaluated on the same independent
items $\mathcal{I}$, so a positive gain means that the selected checkpoint
outperforms the final checkpoint.
\textbf{(b)} Validation NLL and task accuracy can evolve differently
during training. In the original 75-trajectory grid, NLL-based selection
yields a paired accuracy difference of $-0.76$ pp relative to the final
checkpoint (95\% configuration-bootstrap CI $[-1.32,-0.22]$).
This motivates our central question: how much checkpoint variation can
finite validation data reliably convert into better performance on
independent test examples?
}
    \label{fig:motivation}
\end{figure}

Prior work establishes important parts of this problem. Validation loss can
diverge from downstream generation quality during SFT
\citep{ouyang2022training,zhou2023lima}, motivating checkpoint selection
based on held-out generations rather than likelihood alone
\citep{kaur2025instruct,ruan2025unveiling}. More generally,
finite-sample model selection can overfit the data used to choose among
candidates \citep{cawley2010over}. Together, these results show that
selection signals can disagree and that finite selection data can induce
selection bias. The SFT comparisons above use fixed validation sizes, so they
do not quantify how selection performance changes with validation budget or whether
outperforming validation-loss selection also means outperforming the final
checkpoint.

We study these questions through controlled, paired comparisons in small-data
SFT. For each completed training trajectory, we hold candidate checkpoints
and independent test items fixed while varying nested validation subsets. We
compare generated validation accuracy, an inputs-only checkpoint-agreement
signal, and matched negative log-likelihood (NLL). In the primary
comparisons, selected checkpoints are evaluated only on items excluded from
selection. This design separately
measures (i) improvement from additional validation data, (ii) gain over
matched NLL selection, and (iii) gain over the final checkpoint. A
complementary matched-half analysis measures how much selecting and evaluating
on the same finite items can inflate apparent gains.

These three comparisons give different answers. Across 60 mathematical
trajectories and 19 configurations, increasing validation budget from 32
to 305--313 examples improves independent-test accuracy by $+0.32$ pp for
generated-accuracy selection and $+0.29$ pp for checkpoint agreement, with
both 95\% configuration-bootstrap intervals above zero. At full validation
budget, the two generation-based rules outperform matched NLL, yet their
gains over the final checkpoint remain unresolved. A cross-domain Commonsense
replication on 12 new trajectories shows the same qualitative separation:
increasing the validation budget from 32 to
1,024 questions improves generated-accuracy and checkpoint-agreement selection
by $+0.87$ and $+0.27$ pp, while their gains over the final
checkpoint remain unresolved. A separately frozen larger-pool GSM8K
extension further shows a $+0.40$ pp generated-accuracy budget gain with
1,024 distinct source questions. Together, these results show that benefiting
from additional validation data, outperforming NLL selection, and
outperforming the final checkpoint are distinct empirical claims that require
separate evidence.

Our contributions are threefold:
\begin{itemize}
\item
We formulate SFT checkpoint selection as a finite-information decision
problem and make \emph{validation budget} an explicit part of evaluation.
This distinguishes visible within-trajectory variation from gains that can be
reliably recovered and verified on independent examples.

\item
We develop a controlled framework that varies validation budget while holding
completed trajectories, checkpoints, and test items fixed. It measures
validation-budget improvement, gain over matched NLL, gain over the final
checkpoint, and same-item optimism.

\item
Across the original mathematical analysis and a cross-domain Commonsense
replication on newly trained trajectories, additional validation data can
improve generation-based selection and these rules can outperform matched NLL,
while gains over the final checkpoint remain unresolved. Larger-pool analyses
also support generated-accuracy budget gains. These findings show why the three
claims require separate evidence.
\end{itemize}
\section{Measurement Design}
\label{sec:method}

We treat SFT checkpoint selection as a
\emph{finite-information decision problem}: a rule must choose one checkpoint
using limited validation data. We vary the validation budget while holding
the training trajectory, candidate checkpoints, and independent test items
fixed. This design isolates the value of additional validation data from early
stopping or additional training.

\subsection{Problem Formulation and Selection Signals}
\label{subsec:formulation}

A completed trajectory
$\mathcal C=(c_1,\ldots,c_T)$ contains checkpoints at steps
$t_1<\cdots<t_T$, with $c_T$ the final checkpoint at the assigned training
budget. Let $\mathcal D=\{(x_i,y_i)\}$ denote held-out validation data and
$\mathcal I$ an evaluation pool independent of $\mathcal D$.
For a subset $S_n\subseteq\mathcal D$ containing $n$ validation examples,
a signal $s$ defines the selection rule
\begin{equation}
r_s(S_n)\sim\mathrm{Unif}\!\left(
\operatorname*{arg\,min}_{c\in\mathcal C}
\sigma_s s(c;S_n)\right),
\qquad
\sigma_s\in\{+1,-1\},
\label{eq:rule}
\end{equation}
where $\sigma_s=+1$ for signals that are minimized and
$\sigma_s=-1$ for signals that are maximized. Thus, the rule selects a
checkpoint with the best value of signal $s$. If multiple checkpoints tie exactly, the primary analysis averages over
them uniformly. As a validation-free reference, \textbf{Last} always retains
$c_T$. Every comparison therefore asks which checkpoint to retain from the same
completed training trajectory.

We consider three selection signals that use the same number of validation
examples but differ in the information they access. For \emph{generated
validation accuracy}, we greedily generate and normalize an answer for each
validation prompt and maximize exact-match accuracy against the validation
answers. This signal uses both validation inputs and labels.

\emph{Checkpoint agreement} provides an inputs-only alternative. For a prompt
set $\mathcal J$, let $a_t(x)$ be the answer produced by checkpoint
$c_t$. We define the plurality answer and agreement score as
\begin{equation}
a^*(x)\in\operatorname*{arg\,max}_{a}
\sum_{t=1}^{T}\mathbf{1}[a_t(x)=a],
\qquad
s_{\mathrm{agr}}(c_t;\mathcal J)=
\frac{1}{|\mathcal J|}
\sum_{x\in\mathcal J}
\mathbf{1}[a_t(x)=a^*(x)].
\label{eq:agreement}
\end{equation}
Here, $\mathbf{1}[\cdot]$ denotes the indicator function,
$a^*(x)$ is the plurality answer across checkpoints, and higher
$s_{\mathrm{agr}}$ indicates more frequent agreement with these answers.
We maximize $s_{\mathrm{agr}}$. When $\mathcal J=\mathcal D$, checkpoint
agreement uses only held-out prompts, whereas generated validation accuracy
uses their labels. Plurality ties follow the stored checkpoint order, and
checkpoint-score ties follow \Cref{eq:rule}.

As the likelihood-based baseline, we use teacher-forced token-mean negative
log-likelihood (NLL). At each scored response position, let
$p_{ij}^{c}(v)=p_c(v\mid x_i,y_{i,<j})$, and let
$N_{\rm tok}(S_n)$ denote the number of scored response tokens. Then
\begin{equation}
s_{\mathrm{NLL}}(c;S_n)=
-\frac{\sum_{i\in S_n}\sum_j
\log p_{ij}^{c}(y_{ij})}
{N_{\rm tok}(S_n)}.
\label{eq:tfloss}
\end{equation}
We select the checkpoint with the lowest NLL. Perplexity is monotone in NLL
and therefore induces the same checkpoint ordering apart from numerical ties.
In all comparisons using independent validation data, checkpoint choices are
fixed before evaluation on $\mathcal I$.

\subsection{Measuring Selection Gains}
\label{subsec:gains}

We separate three questions that fixed-budget comparisons can conflate:
whether more validation data improve selection, whether a rule outperforms
matched NLL selection, and whether the selected checkpoint outperforms the
final checkpoint.

Let $A_{\mathcal I}(c)$ denote accuracy in percentage points, and let
$\mathcal D_{\rm eligible}$ denote the validation examples available to all
rules in a given comparison, with
$N_{\rm pool}=|\mathcal D_{\rm eligible}|$.
For any selection rule $r$, define
\begin{equation}
\begin{split}
G_r(n)
&=\mathbb E\,A_{\mathcal I}(r(S_n))
  -A_{\mathcal I}(c_T),\\
H_r(n)
&=G_r(n)-G_{\mathrm{NLL}}(n),\\
D_r
&=G_r(N_{\rm pool})-G_r(32).
\end{split}
\label{eq:selection-gains}
\end{equation}
Here, $G_r(n)$ is the \emph{gain over the final checkpoint},
$H_r(n)$ is the gain over matched NLL selection, which we call
\emph{recovery over matched NLL}, and $D_r$ is the
\emph{validation-budget improvement} from 32 examples to the full pool.
Thus, positive $G_r$ means that rule $r$ outperforms the final checkpoint,
positive $H_r$ means that it outperforms matched NLL selection, and positive
$D_r$ means that increasing the validation budget improves the rule.
Importantly, $H_r>0$ does not imply $G_r>0$: outperforming NLL selection
does not by itself show that the selected checkpoint outperforms the final
checkpoint.

The original-pool analysis uses nested budgets of 32, 64, 128,
256, and the full pool. The cross-domain Commonsense replication compares
32 with 1,024 validation questions on newly trained trajectories, and the
frozen larger-pool GSM8K extension uses 32, 128, 512, and 1,024 questions.
Within each trajectory, we average over exact checkpoint-score ties and fixed
validation subsets or permutations as defined by each protocol. Candidate
checkpoint grids and test items remain fixed across rules.

\subsection{Independent Evaluation and Statistical Inference}
\label{subsec:crossfit}

Selecting and evaluating a checkpoint on the same finite items can inflate
apparent performance relative to new items. Our primary comparisons therefore
select checkpoints using $\mathcal D$ and evaluate the frozen choices on the
independent pool $\mathcal I$. In the cross-domain Commonsense replication,
validation measurements and checkpoint choices are completed before
independent-test evaluation.

We measure this effect, which we call \emph{same-item optimism}, with a
fixed-size matched-half diagnostic. We split the evaluation pool into halves
$S$ and $\bar S$, select $r_S=r(S)$ using only $S$, and evaluate the same
selected checkpoint on both halves:
\begin{equation}
\begin{split}
G_{\rm in}
&=\mathbb E A_S(r_S)-A_S(c_T),\\
G_{\rm out}
&=\mathbb E A_{\bar S}(r_S)-A_{\bar S}(c_T),\\
B_r
&=G_{\rm in}-G_{\rm out}.
\end{split}
\label{eq:same-item-gains}
\end{equation}
Generated-accuracy selection uses the labels in $S$, whereas checkpoint
agreement uses only the prompts in $S$. We average both split directions
and 200 partitions within each trajectory. Because the same selected
checkpoint is evaluated on both halves and the selection-set size is fixed,
positive $B_r$ means that the apparent gain is larger on the items used for
selection than on unseen items. It therefore measures same-item optimism,
not an effect of validation budget.

Checkpoint agreement still uses information from the evaluation prompts
themselves, even though it does not use their labels. It is therefore
transductive, so we apply the same sample-splitting protocol.

Paired accuracy differences are computed per trajectory, while
configurations are the resampling unit for pooled uncertainty. Suppose there
are $K$ configurations, and configuration $g$ contains $n_g$ trajectories
with rule-$r$ gains $G_{r,g,s}$. Bootstrap replicate $b$ resamples $K$
configurations and computes
\begin{equation}
\widehat G_r^{(b)}
=
\frac{
\sum_{j=1}^{K}
\sum_{s=1}^{n_{g_j^{(b)}}}
G_{r,g_j^{(b)},s}
}{
\sum_{j=1}^{K}
n_{g_j^{(b)}}
}.
\label{eq:bootstrap}
\end{equation}
We use 20,000 percentile-bootstrap replicates, keeping seeds from each sampled
configuration together. Repeated validation subsets and matched-half partitions
are averaged within each trajectory before resampling, not treated as
independent observations. These intervals are conditional on the observed
configurations and item pools, so dependence from shared model families or
datasets may remain. Intervals including zero are treated as inconclusive, not
as evidence of equivalence. Separate item-level sensitivity analyses hold frozen checkpoint choices fixed.
For the Commonsense replication, a joint sensitivity analysis resamples
configurations and test items; the mathematical and GSM8K item-level
sensitivities condition on the observed configurations. These analyses are
reported separately from the primary configuration-bootstrap intervals.

The paired-gain, matched-half, and mathematical validation-budget analyses are
post-hoc reanalyses of frozen trajectories. The larger-pool GSM8K protocol was
frozen before inference on the new validation pool. The cross-domain
Commonsense protocol was frozen before training, with checkpoint choices fixed
before independent-test evaluation; this replication uses newly trained
trajectories. No selector or
checkpoint grid is retuned after inspecting outcomes.

\section{Experiments and Analysis}
\label{sec:results}

\paragraph{Models, tasks, and completed trajectories.}
The original primary grid comprises \Ntraj{} completed SFT trajectories
across \Nconfig{} configurations and \Nckpt{} evaluated checkpoints.
Qwen3-1.7B, 4B, 8B, and 14B span model scales \citep{yang2025qwen3};
Qwen2.5-7B, Llama-3-8B, and Mistral-7B broaden model-family coverage
\citep{qwen2025qwen2,grattafiori2024llama,jiang2023mistral7b}. Base/Instruct pairs vary initialization,
while LoRA ranks 8/64 and full fine-tuning of Qwen3-1.7B vary adaptation
capacity \citep{hulora}. Mathematical training uses MetaMathQA subsets
\citep{yu2024metamath}, with evaluation on GSM8K and MATH-500
\citep{cobbe2021training,hendrycks2021measuring}; commonsense training uses a
Commonsense170K subset and eight evaluation tasks
\citep{hu2023llm}. Table~\ref{tab:main} lists all primary
configurations. The grid is nonfactorial and concentrated on small-data
mathematical SFT, so it does not isolate effects of model family, scale, or
adaptation method.

\paragraph{Training and measurement.}
Primary runs optimize masked response-token cross-entropy. Training budgets
are 249, 252, or 756 optimizer steps, with checkpoints saved every 25 steps
and at the final step, yielding 10, 11, or 31 candidates. Most configurations
use three seeds; the Llama MATH arm uses six. A trajectory is included only
when valid task and proxy measurements cover its full candidate grid,
including the final checkpoint. Validation examples are assigned by
original-question group before fine-tuning. The reconstructed training and
validation pools have zero overlap at both the original-question-group and
exact-query levels under the recorded split identifiers; Appendix~\ref{app:lineage}
documents the reconstruction and its historical provenance limits. \SplitSizes{}
Teacher-forced measurement uses a 512-token
window and omits sequences with no unmasked response tokens. Evaluation uses
all 1,319 GSM8K items, all 500 MATH-500 items, and 500 items per commonsense
task. Generations are greedy, and answers are normalized before scoring.
Appendix~\ref{app:setup} gives full training and measurement details.

The analyses use overlapping but distinct populations. Appendix
Table~\ref{tab:study-map} maps each analysis to its protocol and population.
The direct-contrast analysis uses 81 trajectories across 26 configurations
(the original grid plus two completed-decay variants); the matched-half
diagnostic uses the original 75 trajectories across 24 configurations; and the mathematical budget analysis uses 60 trajectories
across 19 configurations with complete per-example scoring records. Because
these populations overlap, we analyze them separately rather than as
independent replications, and the direct contrasts and mathematical budget
curves are post-hoc reanalyses of existing predictions. The cross-domain
Commonsense replication adds 12 newly trained trajectories across four
configurations; the frozen 1,024-question GSM8K larger-pool extension uses 12
existing trajectories across four configurations with training and test
predictions fixed. No selector or checkpoint grid is retuned. Protocols,
populations, and controls appear in the appendix.

\FloatBarrier
\subsection{Validation-Budget Response}
\label{subsec:validation-gains}
\FloatBarrier

The mathematical analysis uses nested validation budgets of 32, 64, 128, 256,
and the full pool. Within each trajectory, we average over 200 fixed
permutations while holding candidate checkpoints and test items fixed. The
full pools contain 305 GSM8K, 313 MATH, and 312 Mistral--MATH examples,
corresponding to 244, 229, and 228 original-question groups; some contain
multiple variants of the same source question. NLL uses the matched 512-token
scoring window, and exact checkpoint-score ties are averaged uniformly.
Appendix~\ref{app:budget-gains} reports full curves and sensitivity analyses.

\begin{table}[!htbp]
\caption{\textbf{Validation-budget effects across primary, cross-domain,
and larger-pool analyses.}
Entries are paired independent-test accuracy differences (pp) with marginal
95\% configuration-bootstrap CIs.
$D_r$ measures improvement from 32 examples to the full validation budget,
$H_r$ measures recovery over same-budget matched NLL, and
$G_r$ measures gain over the final checkpoint.
Panel A reports the original mathematical analysis; Panel B is a
cross-domain Commonsense replication on newly trained trajectories; Panel C uses a larger, separately constructed GSM8K
source-question validation pool.
The populations are analyzed separately and are not pooled.}
\label{tab:budget-main}

\centering
\begingroup
\footnotesize
\renewcommand{\arraystretch}{1.10}
\setlength{\tabcolsep}{3.6pt}

\begin{tabular*}{.96\textwidth}
{@{\extracolsep{\fill}}lrrr@{}}

\toprule

\textbf{Rule}
& \multicolumn{1}{c}{\textbf{Budget gain}}
& \multicolumn{1}{c}{\textbf{vs.\ matched NLL}}
& \multicolumn{1}{c}{\textbf{vs.\ final}}\\[-1pt]

& \multicolumn{1}{c}{$D_r$: Full $-$ 32}
& \multicolumn{1}{c}{$H_r$: Full}
& \multicolumn{1}{c}{$G_r$: Full}\\

\midrule


\multicolumn{4}{@{}l}{
\textbf{A.\ Original mathematical pools}
\hfill
\textit{60 trajectories / 19 configurations}}\\[-1pt]

\multicolumn{4}{@{}l}{
\textit{GSM8K + MATH; full = 305/312/313 examples
from 228--244 source groups}}\\

\addlinespace[2pt]

Generated accuracy
& \estci{+0.32}{+0.10}{+0.56}
& \estci{+0.71}{+0.25}{+1.22}
& \estci{+0.01}{-0.33}{+0.35}\\

Checkpoint agreement
& \estci{+0.29}{+0.11}{+0.50}
& \estci{+0.85}{+0.36}{+1.40}
& \estci{+0.15}{-0.08}{+0.37}\\

Matched NLL
& \estci{+0.05}{-0.05}{+0.14}
& $0$ (reference)
& \estci{-0.70}{-1.37}{-0.09}\\

\addlinespace[3pt]
\midrule


\multicolumn{4}{@{}l}{
\textbf{B.\ Cross-domain Commonsense replication}
\hfill
\textit{12 new trajectories / 4 configurations}}\\[-1pt]

\multicolumn{4}{@{}l}{
\textit{8 Commonsense tasks; full = 1,024 validation questions;
4,000 independent test questions}}\\

\addlinespace[2pt]

Generated accuracy
& \estci{+0.87}{+0.55}{+1.32}
& \estci{+1.05}{+0.47}{+1.39}
& \estci{+0.09}{-0.06}{+0.22}\\

Checkpoint agreement
& \estci{+0.27}{+0.17}{+0.37}
& \estci{+1.08}{+0.50}{+1.50}
& \estci{+0.12}{-0.03}{+0.28}\\

Matched NLL
& \estci{+0.42}{-0.22}{+1.13}
& $0$ (reference)
& \estci{-0.96}{-1.38}{-0.26}\\

\addlinespace[3pt]
\midrule


\multicolumn{4}{@{}l}{
\textbf{C.\ Frozen larger-pool GSM8K extension}
\hfill
\textit{12 trajectories / 4 configurations}}\\[-1pt]

\multicolumn{4}{@{}l}{
\textit{Full = 1,024 distinct source questions;
existing trajectories and test predictions held fixed}}\\

\addlinespace[2pt]

Generated accuracy
& \estci{+0.40}{+0.06}{+0.66}
& \estci{+0.45}{-0.10}{+1.00}
& \estci{+0.35}{-0.36}{+1.34}\\

Checkpoint agreement
& \estci{+0.10}{-0.29}{+0.31}
& \estci{+0.14}{-0.94}{+0.80}
& \estci{+0.03}{-0.28}{+0.39}\\

Matched NLL
& \estci{+0.00}{-0.09}{+0.12}
& $0$ (reference)
& \estci{-0.10}{-0.97}{+1.36}\\

\bottomrule

\end{tabular*}
\endgroup
\end{table}

From 32 to full, independent-test accuracy improves by $+0.32$ pp for
generated-accuracy selection $[+0.10,+0.56]$ and $+0.29$ pp for checkpoint
agreement $[+0.11,+0.50]$. At full budget, the two rules outperform matched
NLL by $+0.71$ pp $[+0.25,+1.22]$ and $+0.85$ pp
$[+0.36,+1.40]$, respectively, while gains over the final checkpoint remain
unresolved: $+0.01$ pp $[-0.33,+0.35]$ and $+0.15$ pp
$[-0.08,+0.37]$ (Table~\ref{tab:budget-main}, Panel A). The budget effect
remains positive when retaining one variant per original question and under a
separate test-question bootstrap
(Appendices~\ref{app:unique-questions} and~\ref{app:item-uncertainty}).

\begin{figure}[!htbp]
\centering
\includegraphics[width=.98\textwidth]{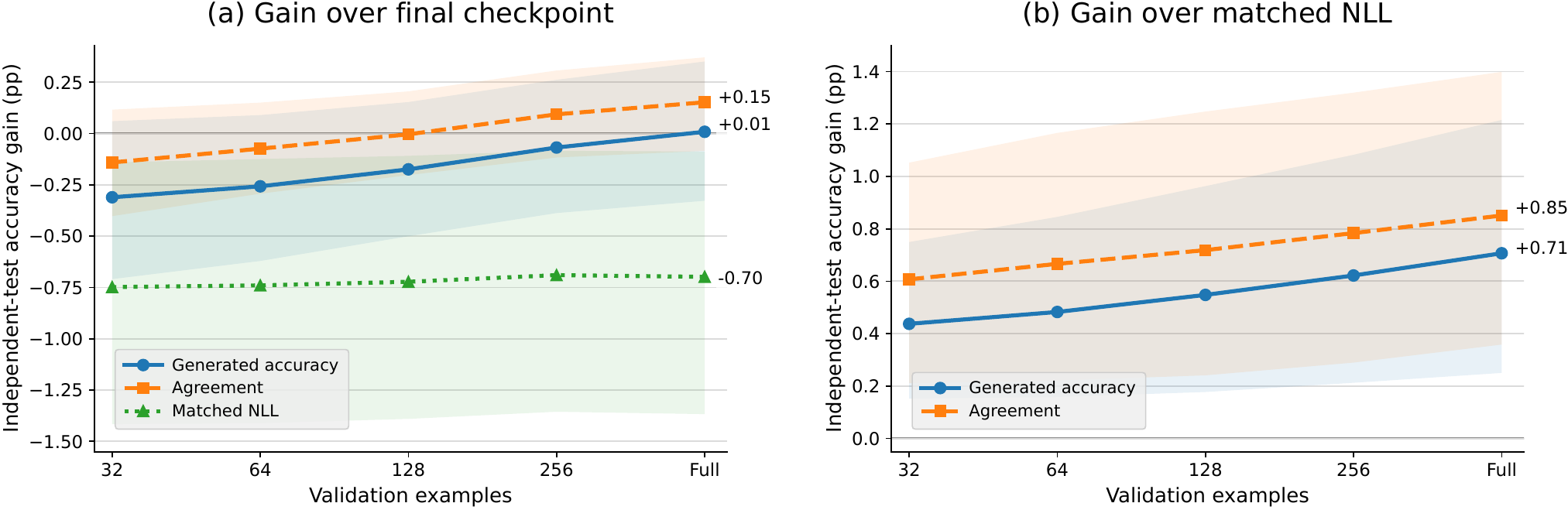}
\caption{\textbf{Validation-budget improvement and gain over final
checkpoint are distinct.}
Post-hoc reanalysis of 60 mathematical trajectories across 19 configurations.
All rules use identical validation subsets and test items. Shading shows
pointwise 95\% configuration-bootstrap intervals. Full denotes 305, 312, or
313 examples by dataset. Positive means higher independent-test accuracy.}
\label{fig:budget-gains}
\end{figure}

A cross-domain Commonsense replication on 12 newly trained trajectories, with
its protocol frozen before training, shows the same separation among the three
claims.
Increasing the budget from
32 to 1,024 questions improves independent-test accuracy by $+0.87$ pp for
generated accuracy and $+0.27$ pp for checkpoint agreement. At full budget,
the two rules outperform matched NLL by $+1.05$ and $+1.08$ pp, while gains
over the final checkpoint remain unresolved ($+0.09$ and $+0.12$ pp;
Table~\ref{tab:budget-main}, Panel B). Thus, validation-budget improvement,
recovery over NLL, and gain over the final checkpoint remain empirically
distinct on new trajectories in a different task domain.

A frozen GSM8K extension with 1,024 distinct source questions yields a
$+0.40$ pp generated-accuracy budget gain, while the agreement contrast
remains unresolved (Table~\ref{tab:budget-main}, Panel C). Full larger-pool
results appear in Appendix~\ref{app:independent-budget}.

\FloatBarrier
\subsection{Recovery over Matched NLL and Gains over the Final Checkpoint}
\label{subsec:norule}
\FloatBarrier

Panel A of Table~\ref{tab:selection-gains} compares two generation-based
rules with matched NLL and the final checkpoint across 81 trajectories and
26 configurations. Generated-accuracy selection and checkpoint agreement
gain $\NovelGenNll$ and $\NovelAgrNll$ pp over matched NLL, respectively;
their gains over the final checkpoint are $-0.01$ and $+0.14$ pp, and both CIs
include zero. Thus, neither rule establishes a gain over the final checkpoint
despite recovering accuracy relative to NLL. Because these comparisons are
paired on the same
trajectories and test items with the same bootstrap resamples, we compare the
paired gains directly rather than relying on separate significance decisions.
For context, NLL
is $-0.76$ pp below the final checkpoint in the original 75-trajectory grid,
with a 95\% configuration-bootstrap CI of $[-1.32,-0.22]$. Full intervals
appear in Table~\ref{tab:selection-gains}; task-level strata are reported in
Appendix~\ref{app:task-strata}.

\begin{table}[!htbp]
\caption{\textbf{Reference gains and same-item optimism.}
\textbf{(A)} Independent-validation selection compares gains over matched NLL and final checkpoint.
\textbf{(B)} Matched-half selection compares selection-half and complementary-half gains at fixed size.
Entries are paired differences (pp) with marginal 95\% configuration-bootstrap intervals; panels use overlapping populations and are analyzed separately.}
\label{tab:selection-gains}
\centering
\begingroup
\footnotesize
\renewcommand{\arraystretch}{1.08}
\setlength{\tabcolsep}{3pt}
\begin{tabular*}{.98\textwidth}{@{\extracolsep{\fill}}lrrr@{}}
\toprule
\multicolumn{2}{@{}l}{\textbf{A. Independent validation selection}}
&\multicolumn{2}{r@{}}{\textit{81 trajectories, 26 configurations}}\\
\midrule
\textbf{Rule} & \textbf{Gain vs. matched NLL}
& \multicolumn{2}{r@{}}{\textbf{Gain vs. final}}\\
Generated accuracy & \estci{+0.74}{+0.38}{+1.14}
&\multicolumn{2}{r@{}}{\estci{-0.01}{-0.27}{+0.24}}\\
Checkpoint agreement & \estci{+0.90}{+0.50}{+1.33}
&\multicolumn{2}{r@{}}{\estci{+0.14}{-0.05}{+0.32}}\\
\midrule
\multicolumn{2}{@{}l}{\textbf{B. Matched-half selection}}
&\multicolumn{2}{r@{}}{\textit{75 trajectories, 24 configurations}}\\
\midrule
\textbf{Rule} & \textbf{Selection-half gain}
& \textbf{Complementary-half gain}
& \textbf{Paired difference}\\
Empirical accuracy maximum & \estci{+1.33}{+0.92}{+1.80}
& \estci{+0.05}{-0.22}{+0.34}
& \estci{+1.29}{+1.01}{+1.59}\\
Checkpoint agreement & \estci{+0.28}{+0.08}{+0.51}
& \estci{-0.03}{-0.21}{+0.17}
& \estci{+0.31}{+0.22}{+0.41}\\
\bottomrule
\end{tabular*}
\endgroup
\end{table}

\FloatBarrier
\subsection{Apparent and Independently Evaluated Gains}
\label{subsec:independent-gains}
\FloatBarrier

On the full evaluation set, the empirically best checkpoint is
$\LastRegret$ pp more accurate than the final checkpoint, but this gap is
descriptive because selection and evaluation use the same finite items.
Panel B of Table~\ref{tab:selection-gains} therefore selects on one half of
the evaluation set and compares the chosen checkpoint with the final
checkpoint on both halves.

Across the original 75 trajectories and 200 partitions, the
accuracy-maximizing rule gains $\NovelOracleIn$ pp on the selection half but
only $\NovelOracleOut$ pp on the complementary half, a paired difference of
$\NovelOracleGap$ pp. The checkpoint-agreement difference is
$\NovelAgrGap$ pp (Table~\ref{tab:selection-gains}, Panel B). These gaps measure
\emph{same-item optimism}: gains can appear larger on selection items than on
unseen items. Agreement is prompt-only and transductive, whereas the
accuracy-maximizing rule uses selection-half labels and is a finite-item
diagnostic, not a population oracle. Because Panels A and B use different
data sizes and distributions, they do not identify a common explanation for
the unresolved gains over the final checkpoint.

\FloatBarrier
\subsection{Scope and Boundary Evidence}
\label{subsec:scope}
\label{sec:validity}
\FloatBarrier

\begin{figure}[!htbp]
\centering
\includegraphics[width=.98\textwidth]{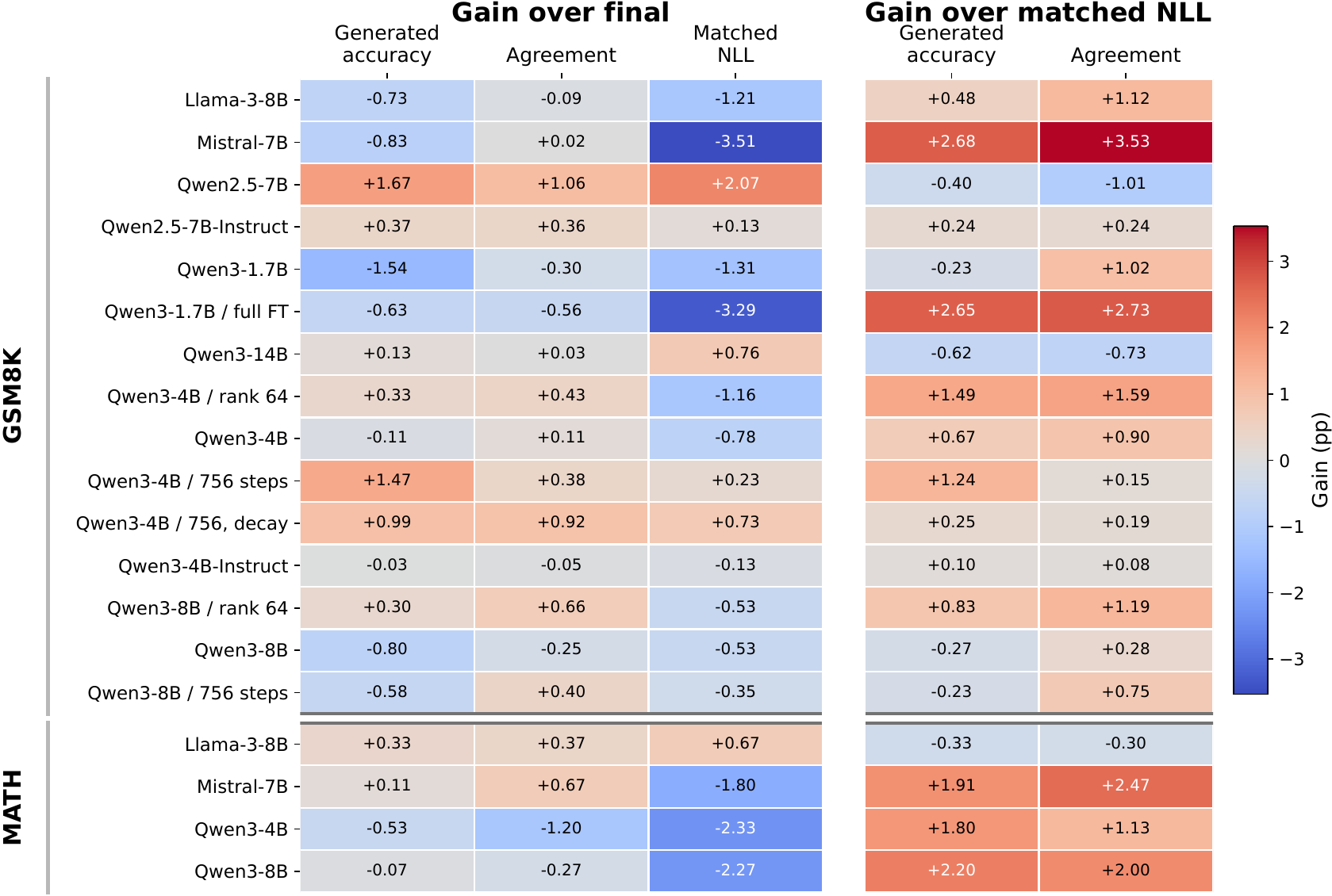}
\caption{\textbf{Full-pool gains vary across mathematical configurations.}
Markers show seed-averaged gains across 60 trajectories in 19 configurations.
These means are descriptive; no subgroup-specific selector is fitted. Panels
show independent-test gains over \textbf{(a)} the final checkpoint and
\textbf{(b)} same-budget matched NLL. Positive values favor the named
selector. Both panels share a horizontal scale; the separator divides GSM8K
and MATH configurations.}
\label{fig:configuration-gains}
\end{figure}

\paragraph{Configuration variation.}
The pooled results hide substantial variation across configurations. At the
full validation budget, generated-accuracy selection has positive mean gains
over the final checkpoint in 9 of 19 configurations and negative gains in 10;
checkpoint agreement is positive in 12 and negative in 7. These signs are
descriptive rather than configuration-level significance tests. Figure~\ref{fig:configuration-gains}
shows this heterogeneity, while Appendix~\ref{app:task-strata} reports descriptive
task-level strata. In the Commonsense replication, generated-accuracy budget
gains are positive in all four configuration means ($+0.45$ to $+1.53$ pp),
and leave-one-configuration-out pooled means remain positive ($+0.65$ to
$+1.01$ pp). A paired sensitivity jointly resampling test items and
configurations keeps the budget-gain and matched-NLL intervals above zero for
both generation-based rules, while both final-checkpoint intervals still cross
zero (Appendix~\ref{app:commonsense-replication}).

Two targeted sensitivities test whether the main three-way pattern depends on
source-question multiplicity or on the realized test questions;
Table~\ref{tab:core-sens} summarizes their point estimates.

\begin{table}[!htbp]
\caption{\textbf{Core-claim sensitivities (pp).} Point estimates for the two
generation-based rules. Full protocols, 95\% intervals, and detailed tables
are reported in Appendices~\ref{app:unique-questions} and~\ref{app:item-uncertainty}.}
\label{tab:core-sens}
\centering
\begingroup
\footnotesize
\renewcommand{\arraystretch}{1.05}
\setlength{\tabcolsep}{3pt}
\begin{tabular*}{.98\textwidth}{@{\extracolsep{\fill}}llrrr@{}}
\toprule
\textbf{Sensitivity} & \textbf{Rule} & \textbf{Budget gain $D_r$} & \textbf{vs. matched NLL $H_r$} & \textbf{vs. final $G_r$}\\
\midrule
One variant / source & Generated accuracy & $+0.22$ & $+0.61$ & $-0.09$\\
& Agreement & $+0.22$ & $+0.78$ & $+0.08$\\
\addlinespace[1pt]
Test-question resampling & Generated accuracy & $+0.32$ & $+0.71$ & $+0.01$\\
& Agreement & $+0.29$ & $+0.85$ & $+0.15$\\
\bottomrule
\end{tabular*}
\endgroup
\end{table}

The sensitivities in Table~\ref{tab:core-sens} preserve the three-way pattern
in Table~\ref{tab:budget-main}: their corresponding 95\% intervals keep $D_r$
and $H_r$ above zero while $G_r$ still crosses zero.

\paragraph{Minimal robustness controls.}
Across all 60 mathematical trajectories, matched 512-token NLL and
complete-response NLL select the same checkpoint, so the truncation used for
the matched baseline does not affect checkpoint selection in this population
(Appendix~\ref{app:answer_audit}). Checkpoint thinning likewise preserves the
qualitative NLL-versus-final result across substantially sparser candidate
grids (Appendix~\ref{app:robustness}). These checks address two direct
measurement concerns without expanding the analysis into a search for a
mechanism behind NLL behavior.

These analyses bound the pooled claims without changing their qualitative
interpretation. Equal validation counts also need not imply equal annotation or
inference cost.

\section{Related Work}
\label{sec:related}

\paragraph{SFT checkpoint selection and finite-sample evidence.}
Validation loss can diverge from downstream generation quality during SFT.
Prior work therefore uses reward-model or generation-based alternatives to
loss-based selection \citep{ouyang2022training,zhou2023lima,kaur2025instruct}.
Validation-perplexity and validation-accuracy selection can also show different
ID/OOD tradeoffs \citep{ruan2025unveiling}, and broader SFT sweeps ask whether
validation loss reliably ranks downstream quality \citep{o2026post}.
More generally, early stopping and finite-sample model selection raise
selection-bias and validation-uncertainty concerns
\citep{prechelt1998automatic,cawley2010over,savvides2023model,mori2026overvalidation}; recent
work also studies stability-aware checkpoint selection under evaluation
uncertainty \citep{xu2026robust}. Relative to prior fixed-budget SFT
comparisons, we vary validation budget while holding trajectories, candidate
checkpoints, and test items fixed and evaluate choices on independent items.
Matched-half, original-question, test-question, and new-pool analyses probe
same-item optimism and robustness, with uncertainty reporting motivated by
work on statistical power and reproducibility
\citep{card2020little,hochlehnert2025sober}.

\paragraph{Broader selection objectives and adjacent methods.}
Likelihood and downstream quality have also been compared in evaluation and
scaling \citep{fang2025wrong,gadre2025language}, including smaller-classifier
evidence favoring loss-based validation \citep{apicella2026don}. Related
selection questions arise in pretraining adaptability, preference alignment,
and reinforcement fine-tuning
\citep{munn2025bayesian,li2026select,li2026getting,kang2026quagmires,nguyen2025uncertainty},
but we study which SFT checkpoint to retain. Agreement-based performance
prediction \citep{baek2022agreement} is closely related to our agreement
signal; verifier-based process supervision \citep{lightman2024let}, weight
averaging \citep{izmailov2018averaging,wortsman2022model}, and self-consistency
\citep{wangself} provide adjacent objectives
or controls rather than checkpoint-selection methods.

\section{Conclusion}
\label{sec:conclusion}

Checkpoint selection under finite validation data is a model-selection problem
with estimation uncertainty; evaluating it therefore requires more than a
fixed-budget comparison between selectors. Across mathematical SFT
trajectories, more validation data improve
generation-based selection and these rules outperform same-budget matched NLL,
but gains over simply retaining the final checkpoint remain unresolved; the
cross-domain Commonsense replication preserves the same qualitative separation.
Matched-half analysis further shows that reusing selection items can overstate
apparent gains. Taken together, these results refine a common intuition:
recovering performance relative to a weaker selector does not imply that finite
validation data have identified a checkpoint that is better than the final
checkpoint on independent examples.

Practically, this motivates reporting validation-budget response, comparison
with a matched-budget baseline, and comparison with the final checkpoint as
distinct quantities, with checkpoint choices frozen before independent
evaluation and uncertainty reported at the appropriate unit. In the settings
studied here, the observed ``best checkpoint'' should therefore be treated as
a data-dependent estimate, not an automatically reliable target. Our evidence
does not establish a universal selector or sufficient validation size; it
provides a protocol for making checkpoint-selection claims more precisely.

\label{maintextend}

\section*{Reproducibility Statement}
Sections~2--3 and Appendices~A--D document the selection rules, checkpoint
 grids, validation budgets, uncertainty procedures, training and measurement
 protocols, and evidence status of the reported analyses. The accompanying
 artifact includes frozen prediction and scoring records, validation
 subset or partition plans, checkpoint-choice records, analysis code, and
 verification scripts for the reported tables and figures.
 Appendix~\ref{app:commonsense-replication} documents the cross-domain
 Commonsense replication, while Appendix~\ref{app:independent-budget}
 documents the frozen larger-pool GSM8K analysis.

\section*{Use of Large Language Models}
Generative AI tools were used for methodology and experimental-design
 feedback, analysis-code development, interpretation and presentation of
 results, manuscript editing, and figure preparation. All AI-assisted work was
 reviewed and verified by the authors, who take responsibility for the final
 content.

\bibliographystyle{iclr2027_conference}
\bibliography{ref-sft}

\clearpage
\appendix
\etocdepthtag.toc{appendix}

\vspace{2em}
\begin{center}
    {\Large\textbf{Appendix}}
\end{center}
\vspace{2em}

\etocsettagdepth{main}{none}
\etocsettagdepth{appendix}{subsection}
\tableofcontents
\clearpage

\section{Supporting Analyses}
\label{app:core}

The appendix reports analyses that directly support the three main claims,
quantify information-reuse bias, or address a concrete measurement concern.
Table~\ref{tab:study-map} summarizes the evidence hierarchy and the population
used by each analysis.

\begin{table}[t]
\caption{\textbf{Analyses of selection gain.}
The populations overlap, so rows are not independent replications.
The first three analyses are post-hoc reanalyses.
The GSM8K new-pool extension freezes its protocol before inference on the new
validation pool; the Commonsense replication uses newly trained trajectories,
freezes its protocol before training, and freezes checkpoint choices before
independent-test evaluation.}
\label{tab:study-map}
\centering
\begingroup
\footnotesize
\renewcommand{\arraystretch}{1.14}
\setlength{\tabcolsep}{4pt}

\begin{tabular}{@{}
>{\raggedright\arraybackslash}p{0.20\textwidth}
>{\raggedright\arraybackslash}p{0.12\textwidth}
>{\raggedright\arraybackslash}p{0.36\textwidth}
>{\raggedright\arraybackslash}p{0.24\textwidth}
@{}}
\toprule
\textbf{Question}
& \textbf{Traj. / config.}
& \textbf{Selection and scoring}
& \textbf{Direct contrast}
\\
\midrule

Recovery over NLL vs.\ gain over final
& 81 / 26
& Select on original validation; score on independent test items.
& Gain over matched NLL and gain over the final checkpoint.
\\

Same-item vs.\ unseen-item gain
& 75 / 24
& Select once on half the evaluation items; score that choice on both halves.
& Selection-half minus complementary-half gain; fixed selection size.
\\

Validation-budget response
& 60 / 19
& Select on shared nested validation subsets; score on a fixed independent test.
& Full-minus-32 gain ($D_r$); same-budget comparisons.
\\

Frozen GSM8K new-pool extension
& 12 / 4
& Select on 32--1,024 newly constructed source questions; reuse existing
training trajectories and test predictions.
& 1,024-minus-32 gain ($D_r$); gains over the final checkpoint and matched NLL.
\\

Cross-domain Commonsense replication
& 12 / 4
& Train 12 new trajectories; select on 32 or 1,024 questions across eight
Commonsense tasks and score frozen choices on 4,000 independent test questions.
& 1,024-minus-32 gain ($D_r$); gains over the final checkpoint and matched NLL.
\\

\bottomrule
\end{tabular}
\endgroup
\end{table}

\FloatBarrier
\subsection{Task-Level Strata}
\label{app:task-strata}

\begin{table}[!ht]
\caption{\textbf{Independent selection gains by task group (pp).}
Results use the 81-trajectory, 26-configuration population of
Table~\ref{tab:selection-gains}A; positive values indicate gain over the final
checkpoint. Task strata and marginal 95\% configuration-bootstrap intervals are
descriptive, not independent confirmations.}
\label{tab:task-gains}
\centering
\begingroup
\footnotesize
\renewcommand{\arraystretch}{1.12}
\setlength{\tabcolsep}{4pt}
\begin{tabular*}{\textwidth}{@{\extracolsep{\fill}}lrrr@{}}
\toprule
\textbf{Rule} & \textbf{GSM8K (48 traj.)} & \textbf{MATH (15)} & \textbf{Commonsense (18)}\\
\midrule
Original NLL & \estci{-0.59}{-1.28}{+0.05} & \estci{-1.01}{-2.30}{+0.25} & \estci{-0.97}{-1.80}{-0.29}\\
Generated accuracy & \estci{-0.01}{-0.42}{+0.41} & \estci{+0.04}{-0.37}{+0.28} & \estci{-0.06}{-0.21}{+0.10}\\
Agreement & \estci{+0.25}{+0.03}{+0.47} & \estci{-0.01}{-0.73}{+0.47} & \estci{-0.02}{-0.13}{+0.13}\\
\bottomrule
\end{tabular*}
\endgroup
\end{table}

Table~\ref{tab:task-gains} reports descriptive task-level gains for the
81-trajectory direct-contrast population. These strata show that the pooled
result is not a substitute for per-task behavior; because the configuration
grid is nonfactorial and subgroup counts are small, they are not interpreted
as independent replications or causal effects of task or model family.

\FloatBarrier
\subsection{Matched-Half Same-Item Optimism}
\label{app:matched-gains}

Main Table~\ref{tab:selection-gains}B uses the original 75 trajectories across
24 configurations and 200 fixed partitions. For each selection half $S$ and
its disjoint complement $\bar S$, the rule selects once on $S$ and the same
checkpoint (or exact uniform-tie expectation) is evaluated on both halves.
The two split directions are averaged before the 200 partitions are averaged
within each trajectory. The halves contain 659/660 GSM8K items, 250 MATH
items, or 2,000 commonsense items. Agreement uses only selection-half prompts;
the empirical accuracy maximum uses selection-half labels. Neither accesses
complementary-half labels when selecting.

Uncertainty uses 20,000 shared configuration-bootstrap draws, retaining all
seeds within each sampled configuration. Partitions are averaged within each
trajectory and are not treated as independent observations. Positive
$B_r=G_{\rm in}-G_{\rm out}$ therefore measures how much larger the apparent
gain is on the finite items used for selection than on complementary items at
the same selection size. It is a diagnostic of information reuse, not an
independent replication or a claim about a population oracle.

\FloatBarrier
\subsection{Complete-Response NLL Control}
\label{app:answer_audit}

The matched NLL baseline scores response tokens within a 512-token window. To
check whether this truncation drives the result, we recompute NLL over the
complete response on the same 60 mathematical trajectories while holding the
validation examples and checkpoint grids fixed. Matched 512-token and
complete-response NLL select the same checkpoint in every trajectory and have
the same paired gain relative to the final checkpoint. Thus, the 512-token
window does not affect NLL checkpoint selection in this population.

\FloatBarrier

\FloatBarrier
\subsection{Validation Budget and Independent Selection Gains}
\label{app:budget-gains}

This post-hoc reanalysis compares generated accuracy, agreement, and matched
512-token NLL on the same 60 mathematical trajectories across 19
configurations. It reuses frozen trajectories and checkpoint grids; no new
training, inference, candidate-grid tuning, or benchmark is introduced.

\paragraph{Common items and original permutations.}
Validation-generation and per-example NLL records are aligned by source
identity. The GSM8K pool has 305 examples from 244 original-question groups;
MATH has 313 from 229 groups. One Mistral--MATH record lacks matched-NLL
measurements in all three seeds, so all three rules use 312 common examples
from 228 groups on those trajectories; all other items are retained. Budget
counts examples, not distinct source problems.

The original 200 deterministic permutations define nested subsets of 32, 64,
128, 256, or the full common pool for every rule. These repetitions probe
subsampling of a fixed pool rather than independent validation datasets.

\paragraph{Selection and outcomes.}
NLL is the sum of stored token losses divided by the sum of scored tokens in
the subset; token counts are verified checkpoint-invariant. The stored
per-example matched-512 scores are numerically distinct from original NLL,
so the curve is labeled matched NLL. Accuracy maximizes correct counts.
Agreement maximizes matches to each item's plurality across the fixed
checkpoint grid, preserving checkpoint-order plurality ties. Item-local
pluralities can be computed before subsetting without accessing other items.
All checkpoint-score ties are evaluated by exact uniform expectation. Selected
checkpoint sets are frozen before their independent-test utility is computed.

For rule $r$, nested subset $S_n$, fixed test set $\mathcal I$, and full
eligible size $N_{\rm pool}$, we report
\begin{equation}
\begin{split}
G_r(n)&=\mathbb E A_{\mathcal I}(r(S_n))-A_{\mathcal I}(c_T),\\
H_r(n)&=G_r(n)-G_{\mathrm{NLL}}(n),\qquad
D_r=G_r(N_{\rm pool})-G_r(32).
\end{split}
\label{eq:budget-gains}
\end{equation}
Expectations average exact checkpoint ties and the 200 permutations within
each trajectory. Positive values indicate higher independent-test accuracy. We
then use 20,000 shared configuration-bootstrap draws (seed 20260825), retaining
all seeds and the trajectory-weighted estimator in \Cref{eq:bootstrap}. Intervals are marginal; curve intervals are pointwise and unadjusted. All are conditional on the fixed
validation and test pools. Because full-pool size differs by dataset, Full is treated as a categorical endpoint.
Equal example budgets do not imply equal token, generation, or annotation cost.

\begin{table}[!ht]
\caption{\textbf{Paired effects of increasing the validation budget (pp).} $G_r(n)$ is independent-test gain over the final checkpoint; $D_r=G_r(\mathrm{full})-G_r(32)$. Brackets are marginal 95\% configuration-bootstrap intervals; repetitions are averaged within each trajectory before resampling. Full pools contain 305--313 examples. Task strata are descriptive, especially MATH with four configurations.}
\label{tab:budget-endpoints}
\centering
\begingroup
\footnotesize
\renewcommand{\arraystretch}{1.12}
\setlength{\tabcolsep}{3pt}
\begin{tabular*}{\textwidth}{@{\extracolsep{\fill}}lrrr@{}}
\toprule
\textbf{Rule} & $G_r(32)$ & $G_r(\mathrm{full})$ & $D_r$\\
\midrule
\multicolumn{4}{@{}l}{\textbf{Pooled} (60 trajectories, 19 configurations)}\\
Generated accuracy & \estci{-0.31}{-0.71}{+0.06} & \estci{+0.01}{-0.33}{+0.35} & \estci{+0.32}{+0.10}{+0.56}\\
Agreement & \estci{-0.14}{-0.40}{+0.12} & \estci{+0.15}{-0.08}{+0.37} & \estci{+0.29}{+0.11}{+0.50}\\
Matched NLL & \estci{-0.75}{-1.42}{-0.14} & \estci{-0.70}{-1.37}{-0.09} & \estci{+0.05}{-0.05}{+0.14}\\
\midrule
\multicolumn{4}{@{}l}{\textbf{GSM8K} (45 trajectories, 15 configurations)}\\
Generated accuracy & \estci{-0.29}{-0.73}{+0.14} & \estci{0.00}{-0.43}{+0.44} & \estci{+0.29}{+0.06}{+0.52}\\
Agreement & \estci{-0.07}{-0.35}{+0.22} & \estci{+0.21}{-0.01}{+0.43} & \estci{+0.28}{+0.13}{+0.43}\\
Matched NLL & \estci{-0.70}{-1.46}{+0.02} & \estci{-0.59}{-1.32}{+0.09} & \estci{+0.10}{+0.03}{+0.19}\\
\midrule
\multicolumn{4}{@{}l}{\textbf{MATH} (15 trajectories, 4 configurations)}\\
Generated accuracy & \estci{-0.38}{-1.13}{+0.31} & \estci{+0.04}{-0.37}{+0.28} & \estci{+0.42}{-0.07}{+1.16}\\
Agreement & \estci{-0.36}{-0.85}{+0.16} & \estci{-0.01}{-0.73}{+0.47} & \estci{+0.34}{-0.20}{+1.05}\\
Matched NLL & \estci{-0.90}{-2.00}{+0.22} & \estci{-1.01}{-2.30}{+0.25} & \estci{-0.11}{-0.46}{+0.07}\\
\bottomrule
\end{tabular*}
\endgroup
\end{table}

\begin{table}[!ht]
\caption{\textbf{Complete pooled validation-budget curves (pp).} Same common-pool subsets and exact uniform checkpoint-score tie expectation for all rules; 60 trajectories across 19 configurations. $G_r(n)$ is independent-test gain over the final checkpoint; $H_r(n)$ is gain over matched token-mean NLL selected on the same subset. Brackets are pointwise, unadjusted 95\% configuration-bootstrap intervals, conditional on the existing pools and grid. Full is a categorical endpoint (305--313 examples).}
\label{tab:budget-curves}
\centering
\begingroup
\footnotesize
\renewcommand{\arraystretch}{1.12}
\setlength{\tabcolsep}{3pt}
\begin{tabular*}{\textwidth}{@{\extracolsep{\fill}}lrrr@{}}
\toprule
\textbf{Rule} & \textbf{Budget} & $G_r(n)$ & $H_r(n)$\\
\midrule
Generated accuracy & 32 & \estci{-0.31}{-0.71}{+0.06} & \estci{+0.44}{+0.15}{+0.75}\\
 & 64 & \estci{-0.26}{-0.62}{+0.09} & \estci{+0.48}{+0.16}{+0.85}\\
 & 128 & \estci{-0.17}{-0.50}{+0.15} & \estci{+0.55}{+0.18}{+0.96}\\
 & 256 & \estci{-0.07}{-0.39}{+0.26} & \estci{+0.62}{+0.21}{+1.08}\\
 & Full & \estci{+0.01}{-0.33}{+0.35} & \estci{+0.71}{+0.25}{+1.22}\\
\midrule
Agreement & 32 & \estci{-0.14}{-0.40}{+0.12} & \estci{+0.61}{+0.21}{+1.05}\\
 & 64 & \estci{-0.07}{-0.29}{+0.15} & \estci{+0.67}{+0.22}{+1.17}\\
 & 128 & \estci{0.00}{-0.20}{+0.21} & \estci{+0.72}{+0.24}{+1.25}\\
 & 256 & \estci{+0.09}{-0.12}{+0.31} & \estci{+0.78}{+0.29}{+1.32}\\
 & Full & \estci{+0.15}{-0.08}{+0.37} & \estci{+0.85}{+0.36}{+1.40}\\
\midrule
Matched NLL & 32 & \estci{-0.75}{-1.42}{-0.14} & \estci{0.00}{0.00}{0.00}\\
 & 64 & \estci{-0.74}{-1.41}{-0.12} & \estci{0.00}{0.00}{0.00}\\
 & 128 & \estci{-0.72}{-1.39}{-0.11} & \estci{0.00}{0.00}{0.00}\\
 & 256 & \estci{-0.69}{-1.36}{-0.08} & \estci{0.00}{0.00}{0.00}\\
 & Full & \estci{-0.70}{-1.37}{-0.09} & \estci{0.00}{0.00}{0.00}\\
\bottomrule
\end{tabular*}
\endgroup
\end{table}

\paragraph{Heterogeneity and inference limits.}
The pooled curves summarize heterogeneous configuration-level effects already
shown in Figure~\ref{fig:configuration-gains}; task strata in
Table~\ref{tab:budget-endpoints} are descriptive, especially MATH with four
configurations. These curves measure the consequence of adding examples from
the observed pools, not a sufficient validation size, an equivalence margin,
or a causal difference between the two generation-based rules. The separate
larger-pool analysis appears in Appendix~\ref{app:independent-budget}. Detailed
per-configuration outputs and reproduction files are provided in the accompanying
artifact.

\FloatBarrier

\FloatBarrier
\subsection{Cross-Domain Commonsense Replication with Protocol Frozen Before Training}
\label{app:commonsense-replication}

We froze the cross-domain replication protocol before training and checkpoint
choices before independent-test scoring. We trained 12 new trajectories from four
configurations---Qwen3-4B-Base and Llama-3-8B Base with LoRA ranks 8 and 64,
three seeds each---with ten candidate checkpoints per trajectory. Optimization
and checkpointing otherwise follow Appendix~\ref{app:setup}. The validation pool
contains 1,024 questions from eight Commonsense tasks (128 per task), and the
independent test pool contains 4,000 questions (500 per task). The SFT training,
validation, and independent test question sets are mutually disjoint under exact
raw-text, normalized-text, and frozen question-hash checks. Validation
measurements and checkpoint choices were completed before independent-test
evaluation, so no independent-test item influenced checkpoint selection.

At 1,024 validation questions, generated accuracy and checkpoint agreement
exceed matched NLL by $+1.05$ pp $[+0.47,+1.39]$ and $+1.08$ pp
$[+0.50,+1.50]$, respectively, while their gains over the final checkpoint
remain unresolved: $+0.09$ pp $[-0.06,+0.22]$ and $+0.12$ pp
$[-0.03,+0.28]$. Increasing the validation budget from 32 to 1,024 questions
improves the two rules by $+0.87$ pp $[+0.55,+1.32]$ and $+0.27$ pp
$[+0.17,+0.37]$ (Table~\ref{tab:budget-main}, Panel B). These intervals use
20,000 configuration-bootstrap draws and retain all seeds within each sampled
configuration.

The generated-accuracy budget effect is positive in all four configuration
means ($+0.45$ to $+1.53$ pp), and leave-one-configuration-out means remain
positive ($+0.65$ to $+1.01$ pp). Its gain over matched NLL is also positive
in all four configurations ($+0.19$ to $+1.43$ pp). Task-level effects are more
heterogeneous, so we treat them as descriptive rather than evidence of a
universal per-task effect.

A paired test-item sensitivity resamples questions within each of the eight
500-question test tasks while jointly resampling configurations and holding
frozen checkpoint-choice distributions fixed. The resulting 95\% intervals
remain positive for generated-accuracy $D_r$ $[+0.45,+1.37]$ and $H_r$
$[+0.39,+1.68]$, and for agreement $D_r$ $[+0.08,+0.47]$ and $H_r$
$[+0.37,+1.80]$; both $G_r$ intervals still include zero. This audit is
conditional on the fixed validation pool and eight-task mixture and does not
establish broad model-family or domain-level generalization.
\FloatBarrier

\FloatBarrier
\subsection{Larger Source-Question GSM8K Validation Pool}
\label{app:independent-budget}

The original mathematical budget analysis subsamples relatively small pools
that contain augmented variants. We therefore ask whether the generated-
accuracy budget response is also visible with a larger, separately constructed
pool of distinct source questions. This protocol was frozen before generating
predictions on the new validation pool and reuses existing training
trajectories and fixed test predictions.

The extension uses four existing GSM8K configurations---Qwen3-4B-Base,
Qwen3-8B-Base, Qwen2.5-7B, and Llama-3-8B---with three seeds each and the
original 11-checkpoint grid, yielding 12 trajectories.

\paragraph{Pool construction and separation.}
From the cached MetaMathQA corpus, we retain records that map uniquely to
official GSM8K training questions with matching extracted answers. We exclude
matches to the available prior SFT/validation sources and main test sets using
normalized text and a word-5-gram overlap check. Deterministic hashing then
selects 1,024 distinct original questions, with one question per normalized
number-template group. The final audit finds zero exact, normalized, or
thresholded near-overlap with those available sources. This strengthens
validation separation without claiming absence from model pretraining.

\paragraph{Frozen measurements and choices.}
We preserve the original Alpaca-style prompt \citep{alpaca}, greedy
decoding, parser, NLL definition, checkpoint grid, and tie conventions. Each of
200 deterministic nested permutations supplies budgets of 32, 128, 512, and
1,024 to all three rules. All new validation measurements were completed
before checkpoint choices were frozen and the existing primary-test curves
were used for scoring. Test accuracy therefore does not influence validation
decisions, although the extension reuses the same trained trajectories and
GSM8K test set.

\paragraph{Results.}
Panel C of Table~\ref{tab:budget-main} reports the endpoint contrasts, while
Table~\ref{tab:independent-curve} shows all frozen budgets. Generated accuracy's
full-minus-32 gain is positive in three of four configuration means and 10 of
12 trajectories, with a pooled increase of $+0.40$ pp $[+0.06,+0.66]$.
Agreement's pooled increase is $+0.10$ pp $[-0.29,+0.31]$. For both rules, the
full-budget contrasts against the final checkpoint and matched NLL remain
unresolved.
With only four configurations, these intervals remain descriptive; the
extension is not an independent training or test replication.

\begin{table}[!ht]
\caption{\textbf{All frozen budget points in the larger-pool GSM8K extension (pp).} Pooled means over the same 12 trajectories. $G_r(n)$ is gain over the final checkpoint and $H_r(n)$ is gain over same-budget matched NLL. Endpoints and their descriptive uncertainty are reported in the main text; complete marginal intervals are provided in the accompanying artifact. Lines between these points would not identify a sufficient budget or saturation.}
\label{tab:independent-curve}
\centering
\begingroup
\footnotesize
\setlength{\tabcolsep}{5pt}
\renewcommand{\arraystretch}{1.12}
\begin{tabular*}{\textwidth}{@{\extracolsep{\fill}}llrrrr@{}}
\toprule
\textbf{Rule} & \textbf{Reference} & $n=32$ & $128$ & $512$ & $1024$\\
\midrule
Generated accuracy & $G_r(n)$ & -0.06 & +0.06 & +0.27 & +0.35 \\
Generated accuracy & $H_r(n)$ & +0.05 & +0.17 & +0.38 & +0.45 \\
Agreement & $G_r(n)$ & -0.06 & -0.02 & +0.02 & +0.03 \\
Agreement & $H_r(n)$ & +0.04 & +0.08 & +0.13 & +0.14 \\
Matched NLL & $G_r(n)$ & -0.10 & -0.10 & -0.11 & -0.10 \\
Matched NLL & $H_r(n)$ & 0.00 & 0.00 & 0.00 & 0.00 \\
\bottomrule
\end{tabular*}
\endgroup
\end{table}

\paragraph{Frozen-choice test-question sensitivity.}
We apply the paired test-question resampling procedure of
Appendix~\ref{app:item-uncertainty} to all 12 trajectories while holding
checkpoint-choice distributions fixed. Table~\ref{tab:independent-item-sensitivity}
reports configuration and test-question intervals separately. Both views retain
a positive generated-accuracy budget contrast, an unresolved agreement budget
contrast, and unresolved full-budget contrasts against both references. These
are conditional sensitivities, not a combined interval. Detailed frozen
choices, pool audits, and verification files are provided in the accompanying
artifact.

\begin{table}[t]
\caption{\textbf{New-pool endpoints under separate uncertainty sources.}
Twelve trajectories across four configurations; all effects are in pp. Configuration
intervals are descriptive and condition on the test set; test-question
intervals condition on configurations and frozen choices. Neither is a joint
interval or an independent replication.}
\label{tab:independent-item-sensitivity}
\centering
\begingroup
\footnotesize
\renewcommand{\arraystretch}{1.12}
\setlength{\tabcolsep}{4pt}
\begin{tabular*}{\textwidth}{@{\extracolsep{\fill}}llrrr@{}}
\toprule
\textbf{Rule} & \textbf{Contrast} & \textbf{Mean} & \textbf{Config. 95\% interval} & \textbf{Test-question 95\% interval}\\
\midrule
Generated accuracy & $D_r$ & $+0.403$ & $[+0.063,+0.664]$ & $[+0.125,+0.686]$\\
 & $G_r(1{,}024)$ & $+0.347$ & $[-0.360,+1.339]$ & $[-0.060,+0.761]$\\
 & $H_r(1{,}024)$ & $+0.449$ & $[-0.101,+0.998]$ & $[-0.028,+0.932]$\\
\addlinespace[2pt]
Agreement & $D_r$ & $+0.095$ & $[-0.292,+0.309]$ & $[-0.155,+0.350]$\\
 & $G_r(1{,}024)$ & $+0.035$ & $[-0.278,+0.392]$ & $[-0.250,+0.316]$\\
 & $H_r(1{,}024)$ & $+0.136$ & $[-0.941,+0.802]$ & $[-0.360,+0.635]$\\
\bottomrule
\end{tabular*}
\endgroup
\end{table}

\FloatBarrier

\FloatBarrier
\subsection{One Variant per Original Question}
\label{app:unique-questions}

The fixed validation pools contain variants of shared source questions. To
assess sensitivity to this multiplicity, we retain all 60 mathematical
trajectories across 19 configurations. Each of 200 deterministic repetitions
selects one eligible variant per original-question group by a fixed hash, then
orders the groups to form nested budgets of 32, 64, 128, and all groups. All
rules use identical subsets, the original candidate grid, and uniform
checkpoint-score ties. Matched NLL remains total token loss divided by total
scored tokens. Checkpoint choices are saved and hashed before test accuracies
are read. The configuration bootstrap and trajectory weighting match the main
budget analysis.

At the full endpoint, the pools contain 244 GSM8K questions, 229 MATH
questions, or 228 questions in the Mistral--MATH common pool.
Table~\ref{tab:unique-questions} preserves the pooled budget benefit and
advantage over matched NLL without establishing superiority to the final
checkpoint. Because the full endpoint contains fewer records and changes
question weighting, this sensitivity does not isolate a causal effect of
deduplication. The accompanying artifact contains the subset plans, frozen choices,
and verification outputs.

\begin{table}[!ht]
\caption{\textbf{Budget gains with one variant per original question (pp).}
All 60 trajectories across 19 configurations; full is 228--244 unique
questions. Brackets are marginal 95\% configuration-bootstrap intervals
conditional on the existing pools. This is a multiplicity sensitivity
analysis, not an independent validation-pool replication.}
\label{tab:unique-questions}
\centering
\begingroup
\footnotesize
\renewcommand{\arraystretch}{1.12}
\setlength{\tabcolsep}{4pt}
\begin{tabular*}{\textwidth}{@{\extracolsep{\fill}}lrrr@{}}
\toprule
\textbf{Rule} & \textbf{Full $-$ 32} & \textbf{Full vs. final} & \textbf{Full vs. matched NLL}\\
\midrule
Generated accuracy & \estci{+0.22}{+0.07}{+0.39} & \estci{-0.09}{-0.47}{+0.28} & \estci{+0.61}{+0.23}{+1.03}\\
Agreement & \estci{+0.22}{+0.08}{+0.37} & \estci{+0.08}{-0.13}{+0.29} & \estci{+0.78}{+0.29}{+1.32}\\
Matched NLL & \estci{+0.07}{-0.01}{+0.16} & \estci{-0.71}{-1.37}{-0.09} & $0$\\
\bottomrule
\end{tabular*}
\endgroup
\end{table}

\FloatBarrier
\subsection{Frozen-Choice Test-Question Sensitivity}
\label{app:item-uncertainty}

The primary configuration-bootstrap intervals condition on the observed test
items. This separate post-hoc analysis quantifies test-question sampling
sensitivity while holding configurations, validation pools, repeated subsets,
and checkpoint selections fixed. It covers all 60 trajectories across 19
configurations in the budget analysis: 45 GSM8K trajectories evaluated on
1,319 questions and 15 MATH trajectories evaluated on 500 questions. No new
training or inference is performed.

\paragraph{Frozen decisions and paired outcomes.}
For each trajectory, rule, and budget, let $p_t$ be the selection probability
of checkpoint $c_t$, averaging the 200 saved validation subsets and exact
uniform checkpoint-score ties. These probabilities are frozen before loading
per-question test outcomes and are not refitted during resampling. With
$z_{ti}\in\{0,1\}$ denoting correctness, the per-question gain over the final
checkpoint is
\begin{equation}
d_i=100\left(\sum_t p_t z_{ti}-z_{Ti}\right).
\label{eq:item-gain}
\end{equation}
For the other contrasts, we subtract either full-budget matched NLL's fixed
per-question utility or the same rule's 32-example utility. Checkpoint and
question identities are aligned to the saved evaluation records before
resampling.

\paragraph{Conditional uncertainty.}
We draw 20,000 bootstrap samples of questions separately within GSM8K and
MATH, sharing each task's sampled question indices across all configurations,
seeds, rules, budgets, and reference checkpoints. This preserves pairing and
dependence induced by shared evaluation items. Trajectory weights remain
fixed, including pooled task weights 45/60 and 15/60; tasks are not reweighted
by test-set size. Validation repetitions are averaged before resampling and
are not treated as independent replications. Table~\ref{tab:item-uncertainty}
reports marginal percentile intervals alongside the original configuration
intervals. The two interval families condition on different fixed quantities;
neither is selected for being narrower, and they are not combined into a
joint interval.

\begin{table}[!ht]
\caption{\textbf{Separate configuration and test-question uncertainty (pp).} The same fixed-pool budget contrasts on 60 trajectories across 19 configurations. Configuration intervals condition on the saved test questions; test-question intervals hold configurations and validation choices fixed, using paired draws shared across all trajectories within each task. Both use 20,000 draws and are marginal 95\% intervals. They are not combined into a joint interval.}
\label{tab:item-uncertainty}
\centering
\begingroup
\footnotesize
\renewcommand{\arraystretch}{1.12}
\setlength{\tabcolsep}{3pt}
\begin{tabular*}{\textwidth}{@{\extracolsep{\fill}}llrrr@{}}
\toprule
\textbf{Rule} & \textbf{Paired contrast} & \textbf{Mean} & \textbf{Configuration CI} & \textbf{Test-question CI}\\
\midrule
Generated accuracy & Full versus final checkpoint & $+0.01$ & $[-0.33,+0.35]$ & $[-0.27,+0.30]$\\
 & Full versus matched NLL & $+0.71$ & $[+0.25,+1.22]$ & $[+0.40,+1.02]$\\
 & Full minus 32 & $+0.32$ & $[+0.10,+0.56]$ & $[+0.17,+0.48]$\\
\addlinespace[2pt]
Agreement & Full versus final checkpoint & $+0.15$ & $[-0.08,+0.37]$ & $[-0.04,+0.35]$\\
 & Full versus matched NLL & $+0.85$ & $[+0.36,+1.40]$ & $[+0.49,+1.22]$\\
 & Full minus 32 & $+0.29$ & $[+0.11,+0.50]$ & $[+0.15,+0.44]$\\
\addlinespace[2pt]
Matched NLL & Full versus final checkpoint & $-0.70$ & $[-1.37,-0.09]$ & $[-1.10,-0.29]$\\
 & Full versus matched NLL & $0.00$ & $[0.00,0.00]$ & $[0.00,0.00]$\\
 & Full minus 32 & $+0.05$ & $[-0.05,+0.14]$ & $[-0.04,+0.15]$\\
\bottomrule
\end{tabular*}
\endgroup
\end{table}

Both sensitivity views retain positive full-minus-32 gains and positive
contrasts of the generation-based rules against matched NLL, while leaving
full-budget superiority to the final checkpoint unresolved. The two interval
families answer different conditional questions and are not independent
replications or evidence of equivalence. Frozen selection probabilities and
verification outputs are included in the accompanying artifact.

\section{Checkpoint-Grid Sensitivity}
\label{app:robustness}

The main analyses compare checkpoints on the recorded candidate grids. To test
whether the NLL-versus-final result depends on unusually dense checkpointing,
we thin each trajectory while always retaining the final checkpoint. Keeping
every second, third, or fourth checkpoint reduces the mean grid size from
13.84 to 7.60, 5.60, and 4.80, respectively. The paired NLL gain relative to
the final checkpoint remains negative: $-0.84$ pp $[-1.40,-0.32]$,
$-0.77$ pp $[-1.29,-0.29]$, and $-0.57$ pp $[-1.15,-0.04]$, respectively.
Thus, the qualitative comparison does not depend on the densest available
checkpoint grid. This is a sensitivity analysis on the existing trajectories,
not an independent replication.

\FloatBarrier

\section{Training and Measurement Details}
\label{app:setup}

\FloatBarrier
\subsection{Training, Checkpoints, and Evaluation Data}

All 24 primary configurations use response-masked cross-entropy with AdamW
\citep{loshchilov2019decoupledweightdecayregularization}, linear learning-rate decay, and a 5\% warmup
ratio. Peak learning rates are $2\times10^{-4}$ for LoRA and
$2\times10^{-5}$ for full fine-tuning; weight decay is zero, and training uses
BF16 precision. The product of per-device microbatch size and gradient
accumulation is 32 (either $2\times16$ or $1\times32$). LoRA dropout is zero,
adapters cover attention and MLP projections, rank 8 is the default, and rank
64 is the capacity control; full fine-tuning updates all base-model parameters.

Primary seeds are $42,43,44$, with $45,46,47$ added for Llama MATH. Partially
evaluated trajectories are excluded rather than assigned a shorter effective
budget. The checkpoint grids are $\{25,50,\ldots,225,249\}$,
$\{25,50,\ldots,250,252\}$, and $\{25,50,\ldots,750,756\}$. Every retained
checkpoint has both proxy and task measurements; zero supervised-token counts,
missing metrics, or a missing final checkpoint invalidate the trajectory.

\begin{table}[!ht]
\caption{\textbf{Primary SFT configuration grid.}
Unless noted, adaptation is rank-8 LoRA with three seeds. The 24 original-grid
configurations plus two direct-contrast completed-decay variants (marked $^*$)
are shown; the nonfactorial grid is reported for coverage, not causal comparison.}
\label{tab:main}
\centering
\begingroup
\footnotesize
\renewcommand{\arraystretch}{0.98}
\setlength{\tabcolsep}{4.2pt}
\begin{tabular*}{\textwidth}{@{\extracolsep{\fill}}llrr@{}}
\toprule
\textbf{Task group} & \textbf{Model / adaptation} & \textbf{Steps} & \textbf{Seeds} \\
\midrule
GSM8K & Llama-3-8B & 252 & 3 \\
GSM8K & Mistral-7B & 252 & 3 \\
GSM8K & Qwen2.5-7B & 252 & 3 \\
GSM8K & Qwen2.5-7B-Instruct & 252 & 3 \\
GSM8K & Qwen3-1.7B & 252 & 3 \\
GSM8K & Qwen3-1.7B (full fine-tuning) & 252 & 3 \\
GSM8K & Qwen3-14B & 252 & 3 \\
GSM8K & Qwen3-4B ($r=64$) & 252 & 3 \\
GSM8K & Qwen3-4B & 252 & 3 \\
GSM8K & Qwen3-4B & 756 & 3 \\
GSM8K & Qwen3-4B (matched decay) & 756 & 3 \\
GSM8K & Qwen3-4B-Instruct & 252 & 3 \\
GSM8K & Qwen3-8B ($r=64$) & 252 & 3 \\
GSM8K & Qwen3-8B & 252 & 3 \\
GSM8K & Qwen3-8B & 756 & 3 \\
GSM8K & Qwen3-8B (matched decay)$^*$ & 756 & 3 \\
\midrule
MATH & Llama-3-8B & 249 & 6 \\
MATH & Mistral-7B & 249 & 3 \\
MATH & Qwen3-4B & 249 & 3 \\
MATH & Qwen3-8B & 249 & 3 \\
\midrule
Commonsense & Llama-3-8B & 249 & 3 \\
Commonsense & Mistral-7B & 249 & 3 \\
Commonsense & Qwen2.5-7B & 249 & 3 \\
Commonsense & Qwen3-4B & 249 & 3 \\
Commonsense & Qwen3-4B & 756 & 3 \\
Commonsense & Qwen3-4B (matched decay)$^*$ & 756 & 3 \\
\bottomrule
\end{tabular*}
\endgroup
\end{table}

The available source pools contain 3,000 GSM8K-oriented MetaMathQA records,
2,990 filtered MATH-oriented records, and 3,000 commonsense records before
splitting. These are not post-split training sizes; Appendix~\ref{app:lineage}
reports eligible training and validation counts. The eight commonsense
evaluation tasks are ARC-Easy, ARC-Challenge, BoolQ, HellaSwag, OpenBookQA,
PIQA, SocialIQA, and WinoGrande
\citep{clark2018think,clark2019boolq,zellers2019hellaswag,mihaylov2018can,bisk2020piqa,sap2019social,sakaguchi2020winogrande}.
Because each contributes 500 items, pooled commonsense accuracy is an
equal-weight task average and remains one task-group score per trajectory, not
eight independent training runs.

\FloatBarrier
\subsection{Training and Validation Source Records}
\label{app:lineage}

A bounded, read-only audit links the available training settings to the
validation pools. The current training implementation assigns whole
\texttt{original\_question} groups using a deterministic hash with data seed
42 and validation fraction 0.1, then tokenizes the two splits separately. All
variants sharing the exact source-question key are assigned to one side.
Table~\ref{tab:lineage} reconstructs this split from the available source
files. Training counts refer to eligible records, not the number of example
presentations across epochs. Both original-question-group and exact-query
intersections are zero in every reconstructed pool, so the reconstructed
splits contain no train--validation leakage under the recorded identifiers.

\begin{table}[!ht]
\caption{\textbf{Available source-pool reconstruction.} Entries give records /
distinct original-question groups. The final column counts shared groups /
exact queries between reconstructed training and validation. This is not a
historical per-run consumed-record manifest.}
\label{tab:lineage}
\centering
\begingroup
\footnotesize
\renewcommand{\arraystretch}{1.12}
\setlength{\tabcolsep}{4pt}
\begin{tabular*}{\textwidth}{@{\extracolsep{\fill}}lrrrr@{}}
\toprule
\textbf{Source pool} & \textbf{Before split} & \textbf{Training} & \textbf{Validation} & \textbf{Intersection}\\
\midrule
GSM8K & 3,000 / 2,421 & 2,695 / 2,177 & 305 / 244 & 0 / 0\\
MATH & 2,990 / 2,083 & 2,677 / 1,854 & 313 / 229 & 0 / 0\\
Commonsense & 3,000 / 2,992 & 2,683 / 2,676 & 317 / 316 & 0 / 0\\
\bottomrule
\end{tabular*}
\endgroup
\end{table}

Saved checkpoint training arguments and logs corroborate these split settings
for 63 of the 81 analyzed trajectories, and the held-out generation order
matches the reconstructed validation pool for a representative checkpoint in
each trajectory. Complete run-level provenance is unavailable for 18 imported
trajectories, and the historical model-specific Mistral--MATH source was not
recovered. The reconstruction therefore establishes zero overlap under the
recorded identifiers where the source split can be reconstructed; it does not
certify historical membership for every run.

Exact source grouping also does not establish semantic separation between
rewritten questions or absence of base-model pretraining exposure. The
accompanying artifact contains the available saved settings, hashed membership,
and consistency checks underlying this audit.

\FloatBarrier
\subsection{Teacher-Forced Scoring and Evaluation Separation}

After causal shifting and masking, response tokens within the 512-token window
are scored and empty supervised sequences are omitted. Token NLL weights all
retained response positions equally. Appendix~\ref{app:answer_audit} verifies
that complete-response scoring selects the same checkpoints on the 60-trajectory
mathematical population.

The frozen overlap audit compares fine-tuning and evaluation questions by exact
match, normalized match, and character 13-gram Jaccard similarity with a 0.8
threshold. It flags none of 5,819 evaluation questions, while a planted
positive control recovers the intended duplicates. These checks do not rule out
semantic duplicates or base-model pretraining exposure.

\FloatBarrier

\section{Limitations}
\label{app:limitations}

Our experiments cover fixed-budget, greedy-decoded SFT up to 14B over a
nonfactorial model/task grid. The cross-domain Commonsense replication and
larger-pool GSM8K analysis broaden the evidence but do not cover all
architectures, tasks, or training regimes. We therefore interpret the results
as evidence that validation-budget improvement, gain over matched NLL, and gain
over the final checkpoint are distinct claims within these settings, not as a
universal selector ranking or validation-size threshold.

\end{document}